\documentclass[11pt,a4paper]{article}
\usepackage[hyperref]{emnlp2020}
\usepackage{times}
\usepackage{latexsym}

\usepackage[ruled, linesnumbered]{algorithm2e}
\usepackage{graphicx}
\usepackage{amssymb}
\usepackage{amsmath}
\usepackage{booktabs}
\usepackage{amsfonts}
\usepackage{multirow}
\usepackage{makecell}
\usepackage{subfigure}
\usepackage{color}
\usepackage{bm}
\newtheorem{definition}{Definition}

\usepackage{microtype}

\aclfinalcopy 

\title{Adaptive Attentional Network for Few-Shot \\ Knowledge Graph Completion}

\author{
    Jiawei Sheng$^{1,2}$, Shu Guo$^{3}$, Zhenyu Chen$^{4}$, Juwei Yue$^{4}$, \\
    {\bf Lihong Wang}$^{3}$\thanks{\hspace{0.15cm}Corresponding author: Lihong Wang.}, {\bf Tingwen Liu}$^{1,2}$ and {\bf Hongbo Xu}$^{1,2}$ \\  
    $^{1}$ Institute of Information Engineering, Chinese Academy of Sciences, Beijing, China\\
    $^{2}$ School of Cyber Security, University of Chinese Academy of Sciences, Beijing, China\\
    $^{3}$ National Computer Network Emergency Response Technical Team/Coordination
    Center of China \\
    $^{4}$ Beijing Advanced Innovation Center of Big Data and Brain Computing,  Beihang University \\
    {\tt shengjiawei@iie.ac.cn, guoshu@cert.org.cn, wlh@isc.org.cn}
    }

\date{}

\begin{document}
\maketitle

\begin{abstract}
  Few-shot Knowledge Graph (KG) completion is a focus of current research, where each task aims at querying unseen facts of a relation given its few-shot reference entity pairs. Recent attempts solve this problem by learning static representations of entities and references, ignoring their dynamic properties, i.e., entities may exhibit diverse roles within task relations, and references may make different contributions to queries. This work proposes an adaptive attentional network for few-shot KG completion by learning adaptive entity and reference representations. Specifically, entities are modeled by an adaptive neighbor encoder to discern their task-oriented roles, while references are modeled by an adaptive query-aware aggregator to differentiate their contributions. Through the attention mechanism, both entities and references can capture their fine-grained semantic meanings, and thus render more expressive representations. This will be more predictive for knowledge acquisition in the few-shot scenario. Evaluation in link prediction on two public datasets shows that our approach achieves new state-of-the-art results with different few-shot sizes. The source code is available at \url{https://github.com/JiaweiSheng/FAAN}.
\end{abstract}

\section{Introduction}
Knowledge Graphs (KGs) like Freebase~\cite{bollacker2008:Freebase}, NELL~\cite{carlson2010:NELL} and Wikidata~\cite{vrandevcic2014:wikidata} are extremely useful resources for NLP tasks, such as information retrieval~\cite{Liu18:KB4IR}, machine reading~\cite{Yang17:KB4MR}, and relation extraction~\cite{Ren17:KB4RE}. A typical KG is a multi-relational graph, represented as triples of the form $(h, r, t)$, indicating that two entities are connected by relation $r$. Although a KG contains a great number of triples, it is also known to suffer from incompleteness problem. KG completion, which aims at automatically inferring missing facts by examining existing ones, has thus attracted broad attention. A promising approach, namely KG embedding, has been proposed and successfully applied to this task. The key idea is to embed KG components, including entities and relations, into a continuous vector space and make predictions with their embeddings. 

\begin{figure}[!t]
    \centering
        \includegraphics[width=0.48\textwidth]{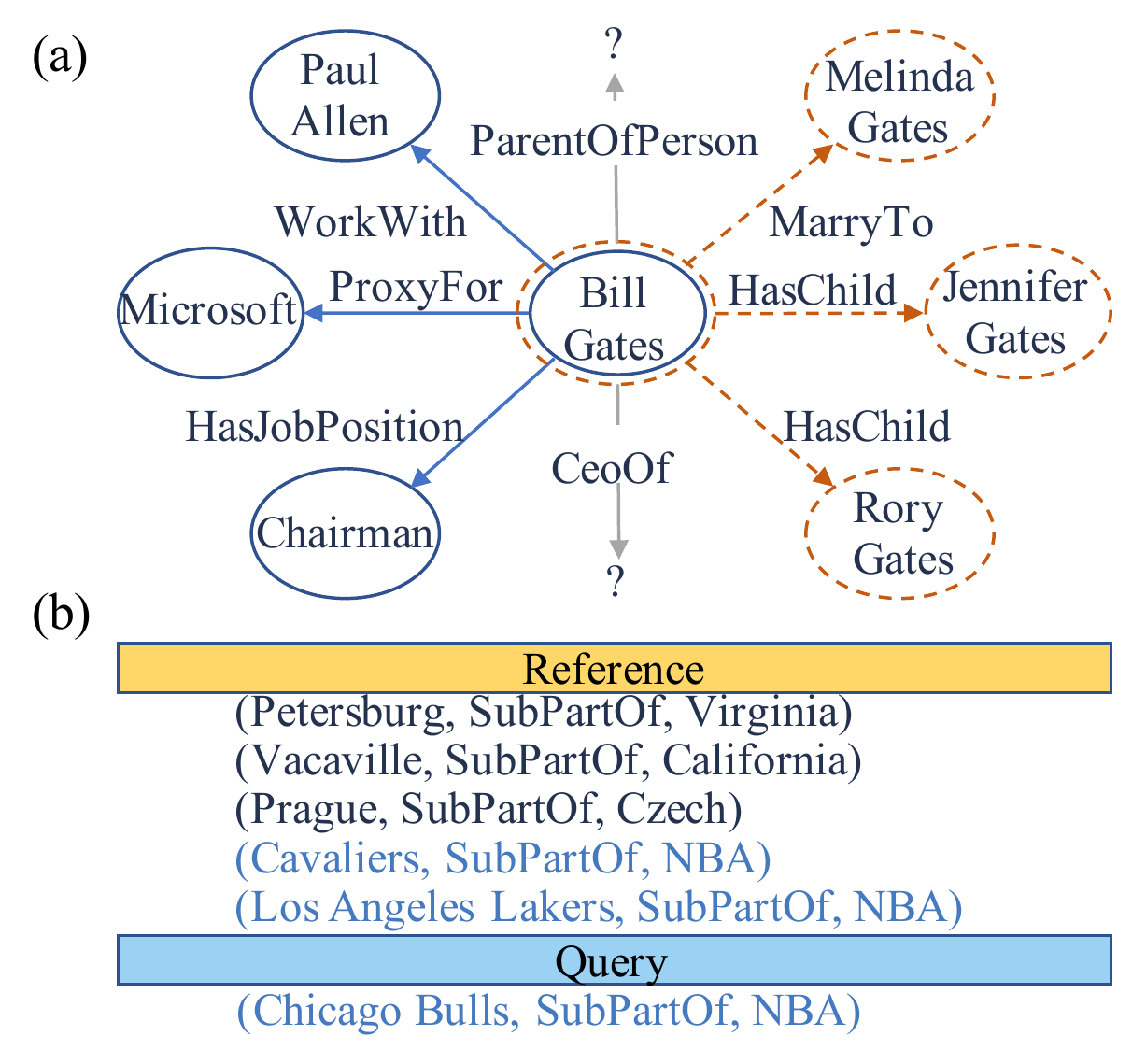}
        \caption{Illustration of dynamic properties in few-shot KG completion: (a) An entity has diverse roles in different tasks; and (b) References show distinct contributions to a particular query.}
        \label{fig:Examples}
\end{figure}

Current KG embedding methods mostly require sufficient training triples for all relations to learn expressive representations ({\it i.e.,} embeddings). In real KGs, a large portion of KG relations is actually long-tail, having only a limited (few-shot) number of relational triples~\cite{Xiong2018:GMatching}. This may lead to low performance of embedding models on KG completion for those long-tail relations.

Recently, several studies~\cite{Chen2019:MetaR,Xiong2018:GMatching,Zhang2020:FSRL} have proposed to address the few-shot issue of KG completion, where one task is to predict tail entity $t$ in a query $(h,r,?)$ given only a few entity pairs of the task relation $r$. These known few-shot entity pairs associated with $r$ are called references. To improve semantical representations of the references, \newcite{Xiong2018:GMatching} and \newcite{Zhang2020:FSRL} devise modules to enhance entity embeddings with their local graph neighbors. The former simply assumes that all neighbors contribute equally to the entity embedding, and in this way the neighbors are always weighted identically. The latter develops the idea by employing an attention mechanism to assign different weights to neighbors, but the weights do not change throughout all task relations. Therefore, both works assign static weights to neighbors, leading to static entity representations when involved in different task relations. We argue that entity neighbors could have varied impacts associated with different task relations. Figure~\ref{fig:Examples}(a) gives an example of head entity $\texttt{\small BillGates}$ associated with two task relations. The left neighbors show his business role, while the right ones show his family role, which reveals quite different meanings. Intuitively, the task relation \texttt{\small CeoOf} is supposed to pay more attention to the business role of entity $\texttt{\small BillGates}$ than the other one.

In addition, task relations can be polysemous, also showing different meanings when involved in different entity pairs. Therefore, the reference triples could also make different contributions to a particular query. Take a task relation \texttt{\small SubPartOf} as an example. As shown in Figure~\ref{fig:Examples}(b), \texttt{\small SubPartOf} associates with different meanings, {\it e.g.}, organization-related as $(\texttt{\small Cavaliers}, \texttt{\small SubPartOf}, \texttt{\small NBA})$ and location-related as $(\texttt{\small Petersburg}, \texttt{\small SubPartOf}, \texttt{\small Virginia})$. Obviously, for query $(\texttt{\small ChicagoBulls}, \texttt{\small SubPartOf}, \texttt{\small ?})$, referring to the organization-related references would be more beneficial. 

To address the above issues, we propose an \underline{A}daptive \underline{A}ttentional \underline{N}etwork for \underline{F}ew-Shot KG completion (FAAN), a novel paradigm that takes dynamic properties into account for both entities and references. Specifically, given a task relation with its reference/query triples, FAAN proposes an adaptive attentional neighbor encoder to model entity representations with one-hop entity neighbors. Unlike the previous neighbor encoder with a fixed attention map in~\cite{Zhang2020:FSRL}, we allow attention scores dynamically adaptive to the task relation under the translation assumption. This will capture the diverse roles of entities through varied impacts of neighbors. Given the enhanced entity representations, FAAN further adopts a stack of Transformer blocks for reference/query triples to capture multi-meanings of the task relation. Then, FAAN obtains a general reference representation by adaptively aggregating the references, further differentiating their contributions to different queries. As such, both entities and references can capture their fine-grained meanings, and render richer representations to be more predictive for knowledge acquisition in the few-shot scenario.

The contributions of this paper are three-fold:

(1) We propose the notion of dynamic properties in few-shot KG completion, which differs from previous paradigms by studying the dynamic nature of entities and references in the few-shot scenario.

(2) We devise a novel adaptive attentional network FAAN to learn dynamic representations. An adaptive neighbor encoder is used to adapt entity representations to different tasks. A Transformer encoder and an attention-based aggregator are used to adapt reference representations to different queries.

(3) We evaluate FAAN in few-shot link prediction on benchmark KGs of NELL and Wikidata. Experimental results reveal that FAAN could achieve new state-of-the-art results with different few-shot sizes.

\section{Related Work}

Recent years have seen increasing interest in learning representations for entities and relations in KGs, {\it a.k.a} KG embedding. Various methods have been devised, and roughly fall into three groups: 1) translation-based models which interpret relations as translating operations between head-tail entity pairs~\cite{bordes2013:TransE,yang2019:TransMS}; 2) simple semantic matching models which compute composite representations over entities and relations using linear mapping operations~\cite{yang2015:DistMult,trouillon2016:ComplEx,liu2017:ANALOGY,sun2019:RotatE}; and 3) (deep) neural network models which obtain composite representations using more complex operations ~\cite{schlichtkrull2017:R-GCN,dettmers2017:ConvE}. Please refer to~\cite{nickel2016:review,wang2017:review, Ji2020:KGEreview} for a thorough review of KG embedding techniques. Traditional embedding models always require sufficient training triples for all relations, thus are limited when solving the few-shot problem. 

Previous few-shot learning studies mainly focus on computer vision~\cite{Sung2018:RelationNetworks}, imitation learning~\cite{Duan2017:OneShotImitation} and sentiment analysis~\cite{Li2019:FewShotSentiment}. Recent attempts~\cite{Xiong2018:GMatching,Chen2019:MetaR,Zhang2020:FSRL} tried to perform few-shot relational learning for long-tail relations. \newcite{Xiong2018:GMatching} proposed a matching network GMatching, which is the first research on one-shot learning for KGs as far as we know. GMatching exploits a neighbor encoder to enhance entity embeddings from their one-hop neighbors, and uses a LSTM matching processor to perform a multi-step matching by a LSTM block. FSRL~\cite{Zhang2020:FSRL} extends GMatching to few-shot cases, further capturing local graph structures with an attention mechanism. \newcite{Chen2019:MetaR} proposed a novel meta relational learning framework MetaR by extracting and transferring shared knowledge across tasks from a few existing facts to incomplete ones. However, previous studies learn static representations of entities or references, ignoring their dynamic properties. This work attempts to learn dynamic entity and reference representations by an adaptive attentional network.

Dynamic properties have also been explored in other contexts outside few-shot relational learning. \newcite{Ji2015:TransD,wang2019:CoKE} performed KG completion by learning dynamic entity and relation representations, but their methods are specially devised for traditional KG completion. \newcite{lu2017:AdaptiveCaption} adopted an adaptive attentional model for image captioning. \newcite{luo2019:AdaptiveRecommendation} tried to model dynamic user preference using a recurrent network with adaptive attention for the sequential recommendation. All these studies demonstrate the capability of modeling dynamic properties to enhance learning algorithms.

\section{Background}\label{sec:background}
Gonsider a KG $\mathcal{G}$ containing a set of triples $\mathcal{T} = \{(h, r, t) \in \mathcal{E} \times \mathcal{R} \times \mathcal{E}\}$, where $\mathcal{E}$ and $\mathcal{R}$ denotes the entity set and relation set, respectively. This work focuses on a challenging link prediction scenario, {\it i.e.}, few-shot KG completion. We follow the standard definition of this task~\cite{Zhang2020:FSRL}:

\begin{definition}[Few-shot KG Completion]\label{def:fewshot} 
    Given a relation $r \in \mathcal{R}$ and its reference set $\mathcal{S}_{r} = \{(h_k, t_k) | (h_k, r, t_k) \in \mathcal{T}\}$, one task is to complete triple $(h, r, t)$ with tail entity $t \in \mathcal{E}$ missing, {\it i.e.}, to predict $t$ from a candidate entity set $\mathcal{C}$ given $(h, r, ?)$. When $|\mathcal{S}_{r}| = K$ and $K$ is very small, the task is called $K$-shot KG completion.  
\end{definition}

\begin{figure*}[ht]
  \centering
  \includegraphics[width=\textwidth]{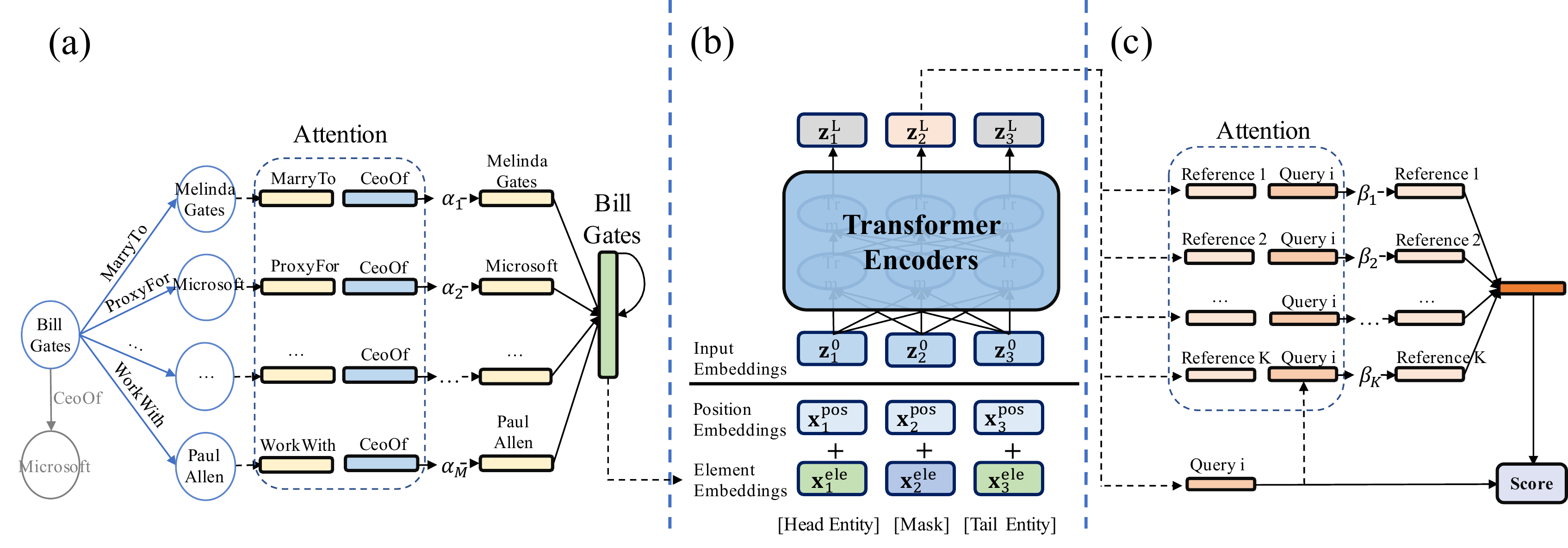}
  \caption{The framework of FAAN: (a) Adaptive neighbor encoder for entities; (b) Transformer encoder for entity pairs; (c) Adaptive matching processor to match $K$-shot references and the query.}
  \label{fig:Model}
\end{figure*}

For this task, the goal of a few-shot learning method is to rank the true tail entity higher than false candidate entities, given few-shot reference entity pairs $\mathcal{S}_{r}$. To imitate such a link prediction task, each training task corresponds to a relation $r \in \mathcal{R}$ with its own reference/query entity pairs, {\it i.e.}, $\mathcal{D}_{r} = \{\mathcal{S}_{r}, \mathcal{Q}_{r}\}$, where $\mathcal{S}_{r}$ only consists of $K$-shot reference entity pairs $(h_{k}, t_{k})$. Additionally, $\mathcal{Q}_{r}=\{(h_m, t_m/\mathcal{C}_{h_m,r})\}$ contains all queries with ground-truth tail entity $t_m$ and the corresponding candidates $\mathcal{C}_{h_m,r}$, where each candidate is an entity in $\mathcal{E}$ selected based on the entity type constraint~\cite{Xiong2018:GMatching}. The few-shot learning method thus could be trained on the task set by ranking the candidates in $\mathcal{C}_{h_m,r}$ given the query $(h_m, r, ?)$ and its references $\mathcal{S}_{r}$. All tasks in training form the {\it meta-training} set, denoted as $\mathcal{T}_{mtr}=\{\mathcal{D}_{r}\}$. 
Here, we only consider a closed set of entities appearing in $\mathcal{E}$.

After suffcient training with meta-training set, the learned model can be used to predict facts of new relation $r'\in \mathcal{R}'$ in testing. The relations used for testing are $unseen$ from meta-training,  {\it i.e.}, $\mathcal{R}' \cup \mathcal{R} = \phi$. Each testing relation $r'$ also has its own few-shot references and queries,  {\it i.e.}, $\mathcal{D}_{r'} = \{\mathcal{S}_{r'}, \mathcal{Q}_{r'}\}$, defined in the same way as in meta-training. All tasks in testing form the {\it meta-testing} set, denoted as $\mathcal{T}_{mte}=\{\mathcal{D}_{r'}\}$. In addition, we also suppose that the model has access to a background KG $\mathcal{G}'$, which is a subset of $\mathcal{G}$ with all the relations excluded from $\mathcal{T}_{mtr}$ and $\mathcal{T}_{mte}$.

\section{Our Approach}
This section introduces our approach FAAN. Given a meta-training set $\mathcal{T}_{mtr}$, the purpose of FAAN is to learn a metric function for predictions by comparing the input query to the given references. To achieve this goal, FAAN consists of three major parts: (1) Adaptive neighbor encoder to learn adaptive entity representations; (2) Transformer encoder to learn relational representations for entity pairs; (3) Adaptive matching processor to compare the query to the given references. Finally, we present the detailed training objective of our model. Figure~\ref{fig:Model} shows the overall framework of FAAN for a task relation $\texttt{\small CeoOf}$.

\subsection{Adaptive Neighbor Encoder for Entities}\label{sec:NbrAgg}
Previous works on embeddings~\cite{schlichtkrull2017:R-GCN,Shang2019:SACN} have demonstrated that explicitly modeling graph contexts benefits KG completion. Recent few-shot relational learning methods encode one-hop neighbors to enhance entity embeddings with equal or fixed attentions~\cite{Xiong2018:GMatching,Zhang2020:FSRL}, ignoring the dynamic properties of entities. To tackle this issue, we devise an adaptive neighbor encoder for entities discerning their entity roles associated with task relations. Specifically, we are given a triple of a few-shot task for relation $r$, {\it e.g.}, $(h,r,t)$. Take the head entity $h$ as a target, and we denote its one-hop neighbors as 
$\mathcal{N}_h=\{(r_{nbr}, e_{nbr})|(h, r_{nbr}, e_{nbr}) \in \mathcal{G}'\}$. Here, $\mathcal{G}'$ is the background KG; $r_{nbr}$, $e_{nbr}$  represent the neighboring relation and entity of $h$ respectively. The aim of the proposed neighbor encoder is to obtain varied entity representations with $\mathcal{N}_h$ to exhibit their different roles when involved in different task relations. Figure~\ref{fig:Model}(a) gives the details of the adaptive neighbor encoder, where $\texttt{\small CeoOf}$ is the few-shot task relation and the other relations such as $\texttt{\small MarryTo}$, $\texttt{\small ProxyFor}$ and $\texttt{\small WorksWith}$ are the neighboring relations of the head entity $\texttt{\small BillGates}$.

As claimed in the introduction, the role of entity $h$ can be varied with respect to the few-shot task relation $r$. However, few-shot task relations are always hard to obtain effective representations by existing embedding models that always require sufficient training data for the relations. Inspired by TransE~\cite{bordes2013:TransE}, we model the task relation embedding $\mathbf{r}$ as a translation between the entity embeddings $\mathbf{h}$ and $\mathbf{t}$, {\it i.e.}, we want $\mathbf{h} + \mathbf{r} \approx \mathbf{t}$ when the triple holds. The intuition here originates from linguistic regularities such as $\texttt{\small Italy} \!-\! \texttt{\small Rome} = \texttt{\small France} - \texttt{\small Paris} $, and such analogy holds because of the certain relation $\texttt{\small CapitalOf}$. Under the translation assumption, we can obtain the embedding of few-shot task relation $r$ given its entity pair $(h,t)$: 
\begin{equation}\label{eq:tran}
    \mathbf{r} = \mathbf{t} - \mathbf{h}
\end{equation} 
where $\mathbf{r}, \mathbf{t}, \mathbf{h} \in \mathbb{R}^{d}$; $\mathbf{t}$ and $\mathbf{h}$ are embeddings pre-trained on $\mathcal{G}'$ with current embedding model such as TransE; $d$ denotes the pre-trained embedding dimension. Actually, the translation mechanism is not the only way to model the task relations. We leave the investigation of other KG embedding methods~\cite{trouillon2016:ComplEx,sun2019:RotatE} to future work. 

Intuitively, relations can reflect roles of an entity. As shown in Figure~\ref{fig:Examples}(a), the task relation $\texttt{\small CeoOf}$ may be more related to $\texttt{\small WorkWith}$ than $\texttt{\small MarryTo}$, since the first two exhibit a business role. That is to say, we can discern the roles of $h$ according to the relevance between the task relation $r$ and the neighboring relation $r_{nbr}$. Hence, we first define a metric function $\psi$ to calculate their relevance score by a bilinear dot product:
\begin{equation}\label{eq:relevance}
    \psi (r, r_{nbr}) = \mathbf{r}^\top \mathbf{W} \mathbf{r}_{nbr} + b
\end{equation}
where  $\mathbf{r}$ and  $\mathbf{r}_{nbr}$ can be obtained by Eq.~(\ref{eq:tran}); both $\mathbf{W} \in \mathbb{R}^{d \times d}$ and $b \in \mathbb{R}$ are learnable parameters. Then, we obtain a role-aware neighbor embedding $\mathbf{c}_{nbr}$ for $h$ by considering its diverse roles:
\begin{align}
    \mathbf{c}_{nbr} &= \sum\nolimits_{e_{nbr}\in \mathcal{N}_h}\alpha_{nbr}\mathbf{e}_{nbr} \label{eq:aggEnt}\\
    \alpha_{nbr} &= \frac{\exp(\psi (r, r_{nbr}))}{\sum\nolimits_{r_{nbr'}\in \mathcal{N}_h}\exp(\psi (r, r_{nbr'}))}\label{eq:nbrAtt}
\end{align}
That means, when neighboring relations are more related to the task relation, $\psi(\cdot, \cdot)$ will be higher and the corresponding neighboring entities would play a more important role in neighbor embeddings.

In order to enhance entity embeddings, we simultaneously couple the pre-trained entity embedding $\mathbf{h}$ and its role-aware neighbor embedding $\mathbf{c}_{nbr}$. Then, $h$ can be formulated as:
\begin{equation}\label{LocalGraphEncode}
    f(h) = \sigma (\mathbf{W}_1 \mathbf{h} + \mathbf{W}_2 \mathbf{c}_{nbr}), 
\end{equation}
where $\sigma(\cdot)$ denotes activation function, and we use Relu; $\mathbf{W}_{1}, \mathbf{W}_{2} \in \mathbb{R}^{d \times d}$ are learnable parameters. Entity representations obtained in this way shall 1) preserve individual properties made by the current embedding model, and 2) possess diverse roles adaptive to different tasks. The above procedure also holds for the candidate tail entity $t$. 

\subsection{Transformer Encoder for Entity Pairs}
Based on enhanced entity embeddings, we are going to derive embeddings of entity pairs. Figure~\ref{fig:Model}(b) gives the details of Transformer encoder for entity pairs. FAAN borrows ideas from recent techniques for learning dynamic KG embeddings~\cite{wang2019:CoKE}. Given an entity pair in a task of $r$, {\it i.e}, $(h,t) \in \mathcal{D}_{r}$, we take each entity pair with its task relation as a sequence $X = (x_{1}, x_{2}, x_{3})$, where the first/last element is head/tail entity, and the middle is the task relation. For each element $x_{i}$ in $X$, we construct its input representation as:
\begin{equation}
    \mathbf{z}_{i}^{\mathrm{0}} = \mathbf{x}_{i}^{\mathrm{ele}} + \mathbf{x}_{i}^{\mathrm{pos}}
\end{equation}
where $\mathbf{x}_{i}^{\mathrm{ele}}$ denotes the element embedding, and $\mathbf{x_{i}}^{\mathrm{pos}}$ the position embedding. Both $\mathbf{x_{1}}^{\mathrm{ele}}$ and $\mathbf{x_{3}}^{\mathrm{ele}}$ are obtained from the adaptive neighbor encoder. We allow a position embedding for each position within length 3. After constructing all input representations, we feed them into a stack of $L$ Transformer blocks~\cite{Vaswani2017:Transformer} to encode $X$ and obtain:
\begin{equation} \label{eq:entpair}
\mathbf{z}_{i}^{l}= \mathrm{Transformer}(\mathbf{z}_{i}^{l-1}), l=1,2,\cdots, L.
\end{equation}
where $\mathbf{z}_{i}^{l}$ is the hidden state of $x_i$ after the $l$-th layer. Transformer adopts a multi-head self-attention mechanism, with each block allowing each element to attend to all elements with different weights in the sequence. 

To perform the few-shot KG completion task, we restrict the mask solely to the task relation $r$ ({\it i.e. $x_{2}$}), so as to obtain meaningful entity pair embeddings. The final hidden state $\mathbf{z}_{2}^{L}$ is taken as the desired representation for the entity pair in $\mathcal{D}_{r}$. Such representation encodes semantic roles of each entity, and thus helps discern fine-grained meanings of task relations associated with different entity pairs. For more details about Transformer, please refer to~\newcite{Vaswani2017:Transformer}.

\subsection{Adaptive Matching Processor}
To make predictions by comparing the query to references, we devise an adaptive matching processor considering different semantic meanings of the task relation. Figure~\ref{fig:Model}(c) gives the details of adaptive matching processor.

In order to compare one query to $K$-shot references, we are going to obtain a general reference representation for the given reference set $\mathcal{S}_r$. Considering the various meanings of the task relation, we define a metric function $\delta (q_{r}, s_{rk})$ that measures the semantic similarity of the query $q_{r}$ and the reference triple $s_{rk}$. For simplicity, we achieve $\delta (q_{r}, s_{rk})$ with simple but effective dot product: 
\begin{equation}\label{eq:delta}
    \delta \left(q_{r}, s_{rk}\right) = \mathbf{q}_r\cdot \mathbf{s}_{rk}
\end{equation}
Unlike current few-shot relational learning models that learn static representations when predicting different queries, we adopt attention mechanism to obtain a general reference representation $g\left(\mathcal{S}_r\right)$ adaptive to the query. This can be formulated as:
\begin{align}
    g\left(\mathcal{S}_r\right) &= \sum\nolimits_{s_{rk} \in \mathcal{S}_r} \beta_k \mathbf{s}_{rk} \label{eq:aggRef}\\
    \beta_{k} &= \frac{\exp(\delta \left(q_{r}, s_{rk}\right))}{\sum\nolimits_{s_{rj} \in \mathcal{S}_{r}} \exp(\delta \left(q_{r}, s_{rj}\right))}
\end{align}
Here, $\beta_{k}$ denotes the attention score of a reference; $s_{rk} \triangleq (h_k, t_k) \in \mathcal{S}_{r}$ denotes the $k$-th reference in the task of $r$, and $\mathbf{s}_{rk}$ is its embedding; $\mathbf{q}_r$ is the embedding of a query $q_r$ in $\mathcal{Q}_{r}$. Both $\mathbf{s}_{rk}$ and $\mathbf{q}_r$ are obtained by Eq.~(\ref{eq:entpair}), to capture their fine-grained meanings. Eq.~(\ref{eq:aggRef}) leads to the fact that references having similar meanings to the query would be more referential, making reference set $\mathcal{S}_{r}$ have an adaptive representation to different queries. 

To make predictions, we define a metric function $\phi \left(q_r, \mathcal{S}_{r}\right)$ to measure the semantic similarity of the query $q_r$ and the reference representation $\mathcal{S}_{r}$:
\begin{equation}
    \phi \left(q_r, \mathcal{S}_{r}\right) = \mathbf{q}_r \cdot g\left(\mathcal{S}_r\right). 
\end{equation}
$\phi(\cdot)$ is expected to be large if the query holds, and small otherwise. Here, $\phi \left(\cdot, \cdot \right)$ can also be implemented with alternative metrics such as cosine similarity or Euclidean distance. 

\subsection{Model Training}
With the adaptive neighbor encoder, the Transformer encoder and the adaptive matching processor, the overall model of FAAN is then trained on meta-training set $\mathcal{T}_{mtr}$. $\mathcal{T}_{mtr}$ is obtained by the following way. For each few-shot relation $r$, we randomly sample $K$-shot positive entity pairs from $\mathcal{T}$ as the reference set $\mathcal{S}_r$. The remaining entity pairs are utilized as positive query set $Q_{r}=\{(h_m, t_m)\}$. Then we construct a set of negative queries $Q_{r}^{-}=\{(h_m, t_m^{-})\}$ by randomly corrupting the tail entity of $(h_m, t_m)$, where $t_m^{-}\in\mathcal{E}\setminus\{t_m\}$. Then, the overall loss is formulated as: 
\begin{equation}
    \mathcal{L}=\sum_{r}\!\sum_{q_r\in Q_{r}}\!\sum_{q_{r}^{-}\in Q_{r}^{-}}\!\! \left[\gamma \!+\! \phi(q_{r}^{-}, \mathcal{S}_r) \!-\! \phi(q_{r}, \mathcal{S}_r)\right]_{+}
\end{equation}
where $[x]_{+} = \max(0, x)$ is standard hinge loss, and $\gamma$ is a margin separating positive and negative queries. To minimize $\mathcal{L}$, we take each relation in $\mathcal{T}_{mtr}$ as a task, and adopt a batch sampling based meta-training procedure proposed in~\cite{Zhang2020:FSRL}. To optimize model parameters in $\Theta$ and Transformer, we use Adam optimizer~\cite{Kingma2014:Adam}, and further impose $L_2$ regularization on the parameters to avoid over-fitting.
    
\section{Experiments}
In this section, we conduct link prediction experiments to evaluate the performance of FAAN. 

\subsection{Datasets}
We conduct experiments on two public benchmark datasets: NELL and Wiki\footnote{\url{https://github.com/xwhan/One-shot-Relational-Learning}}. In both datasets, relations that have less than 500 but more than 50 triples are selected to construct few-shot tasks. There are 67 and 183 tasks in NELL and Wiki, respectively. We use original 51/5/11 and 133/16/34 relations in NELL and Wiki, respectively, for training/validation/testing as defined in Section~\ref{sec:background}. Moreover, for each task relation, both datasets also provide candidate entities, which are constructed based on the entity type constraint~\cite{Xiong2018:GMatching}. More details are shown in Table~\ref{tab:Datasets}.
\begin{table}[t]
	\centering\footnotesize\setlength{\tabcolsep}{3pt}
	\begin{tabular*}{0.48 \textwidth}{@{\extracolsep{\fill}}@{}l|rrrr@{}}
		\toprule
		Dateset  & \# Ent. & \# Rel. & \# Triples & \# Tasks \\
		\midrule
		NELL & 68,545 & 358 & 181,109 & 67 \\
		Wiki & 4,838,244 & 822 & 5,859,240 & 183 \\
		\bottomrule
	\end{tabular*}
	\caption{Statistics of datasets. Each column represents the number of entities, relations, triples and tasks.}
	\label{tab:Datasets}
\end{table}

\begin{table*}[t]
    \centering\footnotesize\setlength{\tabcolsep}{5pt}
    \begin{tabular*}{1 \textwidth}{@{\extracolsep{\fill}}@{}lccccccccc@{}}
    \toprule
    \multicolumn{1}{c}{} & \multicolumn{4}{c}{NELL} && \multicolumn{4}{c}{Wiki} \\
    \cmidrule{2-5}
    \cmidrule{7-10}
    \multicolumn{1}{c}{} & MRR & Hits@10 & Hits@5 & Hits@1 && MRR & Hits@10 & Hits@5 & Hits@1 \\
    \midrule
    TransE~\cite{bordes2013:TransE}                 & .174 & .313 & .231 & .101 && .133 & .187 & .157 & .100 \\
    DistMult~\cite{yang2015:DistMult}               & .200 & .311 & .251 & .137 && .071 & .151 & .099 & .024 \\
    ComplEx~\cite{trouillon2016:ComplEx}            & .184 & .297 & .229 & .118 && .080 & .181 & .122 & .032 \\
    SimplE~\cite{Kazemi018:SimplE}                  & .158 & .285 & .226 & .097 && .093 & .180 & .128 & .043 \\
    RotatE~\cite{sun2019:RotatE}                    & .176 & .329 & .247 & .101 && .049 & .090 & .064 & .026 \\
    \midrule
    GMatching (MaxP)~\cite{Xiong2018:GMatching}     & .176 & .294 & .233 & .113 && .263 & .387 & .337 & .197 \\
    GMatching (MeanP)~\cite{Xiong2018:GMatching}    & .141 & .272 & .201 & .080 && .254 & .374 & .314 & .193 \\
    GMatching (Max)~\cite{Xiong2018:GMatching}      & .147 & .244 & .197 & .090 && .245 & .372 & .295 & .185 \\
    FSRL~\cite{Zhang2020:FSRL}                      & .153 & .319 & .212 & .073 && .158 & .287 & .206 & .097 \\ 
    MetaR~\cite{Chen2019:MetaR}                     & .209 & .355 & .280 & .141 && .323 & .418 & .385 & .270 \\
    \midrule
    FAAN (Ours) & \bf .279 & \bf .428 & \bf .364 & \bf .200 && \bf .341 & \bf .463 & \bf .395 & \bf .281 \\ 
    \bottomrule
    \end{tabular*}
    \caption{Results of 5-shot link prediction on NELL and Wiki. {\bf Bold} numbers denote the best results of all methods.}
    \label{tab:MainResults}
\end{table*}

\subsection{Comparision Methods}
In order to evaluate the effectiveness of our method, we compare our method against the following two groups of baselines:

\paragraph{KG embedding method.} 
This kind of method learns entity/relation embeddings by modeling relational structures in KG. We adopt five widely used methods as baselines: TransE~\cite{bordes2013:TransE}, DistMult~\cite{yang2015:DistMult}, ComplEx~\cite{trouillon2016:ComplEx}, SimplE~\cite{Kazemi018:SimplE} and RotatE~\cite{sun2019:RotatE}. All KG embedding methods require sufficient training triples for each relation, and learn static representations of KG.

\paragraph{Few-shot relational learning method.} 
This kind of method achieves state-of-the-art performance of few-shot KG completion on NELL and Wiki datasets. GMatching~\cite{Xiong2018:GMatching} adopts a neighbor encoder and a matching network, but assumes that all neighbors contribute equally. FSRL~\cite{Zhang2020:FSRL} encodes neighbors with a fixed attention mechanism, and applies a recurrent autoencoder to aggregate references. MetaR~\cite{Chen2019:MetaR} makes predictions by transferring shared knowledge from the references to the queries based on a novel optimization strategy. All the above methods learn static representations of entities or references, ignoring their dynamic properties. 

\subsection{Implementation Details}
We perform 5-shot KG completion task for all the methods. Our implementation for KG embedding baselines is based on OpenKE\footnote{\url{https://github.com/thunlp/OpenKE/tree/OpenKE-PyTorch}}~\cite{Han2018:OpenKE} with their best hyperparameters reported in the original literature. During training, all triples in background KG $\mathcal{G}'$ and training set, as well as few-shot reference triples of validation and testing set are used to train models. For few-shot relational learning baselines, we extend GMatching from original one-shot scenario to few-shot scenario by three settings: obtaining general reference representation by mean/max pooling (denoted as MeanP/MaxP) over references, or taking the reference that leads to the maximal similarity score to the query (denoted as Max). Because FSRL was reported in completely different experimental settings, we reimplement the model to make a fair comparison. We directly report the original results of MetaR with pre-trained embeddings to avoid re-implementation bias.

For all implemented few-shot learning methods, we initialize entity embeddings by TransE. The entity neighbors are randomly sampled and fixed before model training, and the maximum number of neighbors $M$ is fixed to 50 on both datasets. The embedding dimensionality is set to 50 and 100 for NELL and Wiki, respectively. For FAAN, we further set the number of Transformer layers to 3 and 4, and the number of Transformer heads to 4 and 8, respectively. To avoid over-fitting, we also apply dropout to the neighbor encoder and the Transformer layer with the rate tuned in $\{0.1, 0.3\}$. The $L_2$ regularization coefficient is tuned in $\{0, 1e^{-4}\}$. The margin $\gamma$ is fixed to 5.0. The optimal initial learning rate $\eta$ for Adam optimizer is $5e^{-5}$ and $6e^{-5}$ for NELL and Wiki respectively, which is warmed up over the first 10k training steps, and then linearly decayed. We evaluate all methods for every 10k training steps, and select the best models leading to the highest MRR (described later) on the validation set within 300k steps. The optimal hyperparameters are tuned by grid search on the validation set.

\vspace{-8pt}

\subsection{Evaluation Metrics}
To evaluate the performance of all methods, we measure the quality of the ranking of each test triple among all tail substitutions in the candidates: $(h_m, r', t_m')$, $t_k' \in \mathcal{C}_{h_m,r'}$. We report two standard evaluation metrics on both datasets: MRR and Hits@N. MRR is the mean reciprocal rank and Hits@N is the proportion of correct entities ranked in the top $N$, with $N=1, 5, 10$.

\begin{figure}[!t]
	\centering
	\includegraphics[width=0.47\textwidth]{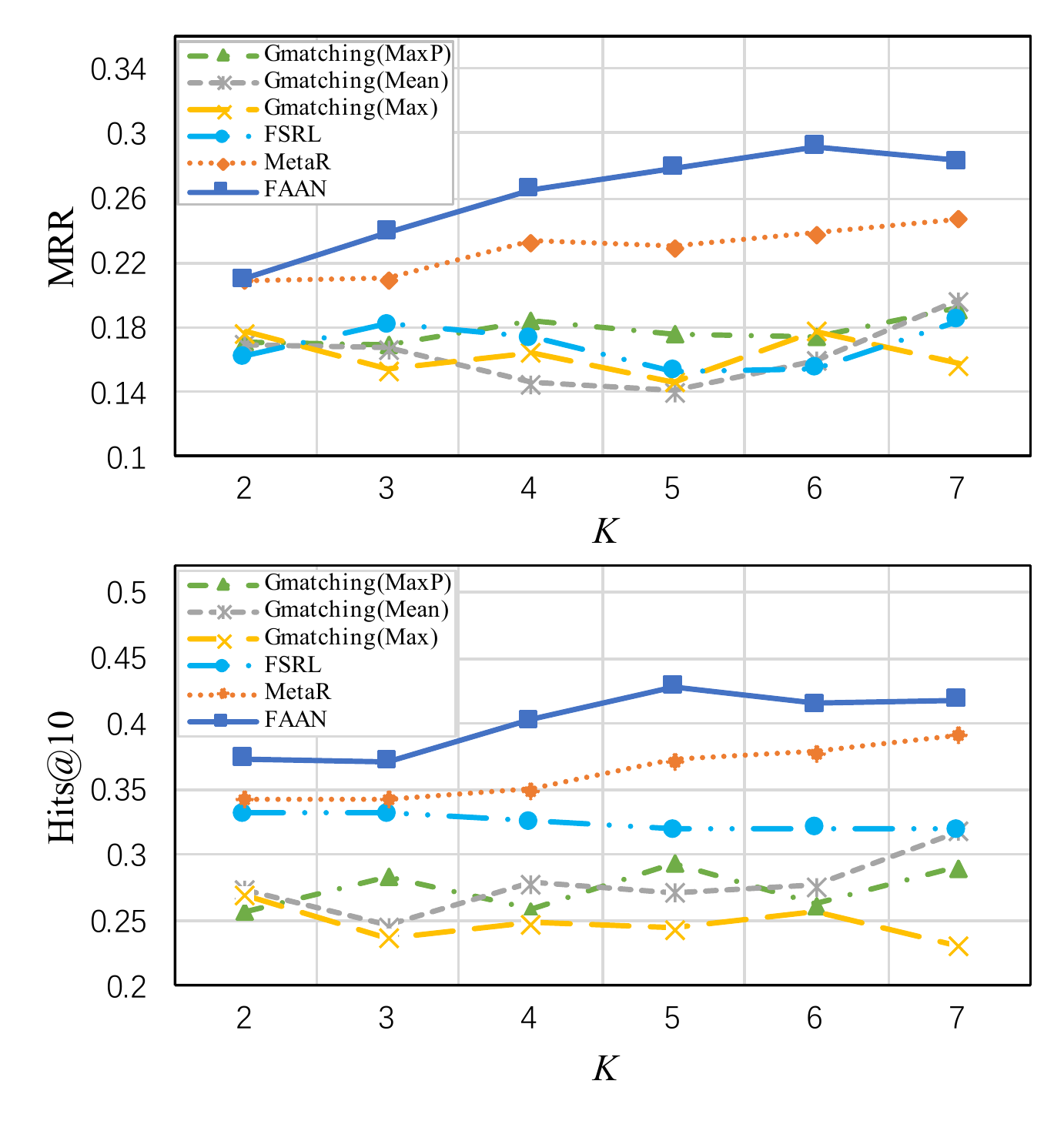}
	\caption{Impact of few-shot size $K$ on NELL dataset.}
	\label{fig:K-shot}
\end{figure}

\subsection{Main Results in Link Prediction}
The performance of all models on NELL and Wiki are shown in Table~\ref{tab:MainResults}. The table reveals that: 

(1) Compared to the traditional KG embedding methods, our model achieves better performance on both datasets. The experimental results indicate that our few-shot learning method is more suitable for solving few-shot issues.
    
(2) Compared to the few-shot learning baselines, our model also consistently outperforms them on both datasets in all metrics. Compared to the best performing baseline MetaR, FAAN achieves an improvement of 33.5\%/20.6\% in MRR/Hits@10 on NELL test data, and an improvement of 5.6\%/10.8\% on Wiki test data, respectively. It demonstrates that exploiting the dynamic properties of KG can indeed improve the performance of few-shot KG completion.

\subsection{Impact of Few-Shot Size}

We conduct experiments to analyze the impact of few-shot size $K$. Figure~\ref{fig:K-shot} reports the performance of models on NELL data in different settings of $K$. The figure shows that: 

(1) Our model outperforms all baselines by a large margin under different $K$, showing the effectiveness of our model in the few-shot scenario.

(2) An interesting observation is that a larger reference set does not always achieve better performance in the few-shot scenario. The reason is probably that few-shot scenario makes the performance sensitive to available references. Take the task relation \texttt{\small SubPartOf} in Figure~\ref{fig:Examples}(b) as an example. When making predictions for organization-related queries, injecting more location-related references is not necessarily useful. Even so, FAAN still gets relatively stable improvements compared to most baselines like GMatching and FSRL. The robustness to few-shot size comes from better reference embeddings generated by the adaptive aggregator.

\begin{table}[!t]
    \centering\footnotesize\setlength{\tabcolsep}{5pt}
    \begin{tabular*}{0.48 \textwidth}{@{\extracolsep{\fill}}@{}lc|cccc@{}}
    \toprule
    Variants && MRR & Hits@10 & Hits@5 & Hits@1 \\
    \midrule
    A1 && .138 & .295 & .169 & .072 \\
    A2 && .209 & .382 & .294 & .120 \\
    A3 && .274 & .411 & .340 & .199 \\
    \midrule
    B1 && .235 & .376 & .301 & .166  \\
    B2 && .271 & .413 & .348 & .195  \\
    \midrule
    C1 && .219 & .355 & .287 & .144  \\
    C2 && .244 & .395 & .317 & .171 \\
    C3 && .212 & .374 & .295 & .122 \\
    \midrule
    Ours &&  \bf .279 & \bf .428 & \bf .364 & \bf .200  \\ 
    \bottomrule
    \end{tabular*}
    \caption{Results of model variants on NELL dataset. {\bf Bold} numbers denote the best results of all variants.}
    \label{tab:ablation}
\end{table}

\subsection{Discussion for Model Variants}

To inspect the effectiveness of the model components, we show results of experiments for model variants in Table~\ref{tab:ablation}:


(A) \textbf{Neighbor Encoder Variants:} In A1, we replace the encoder by mean pooling module used in GMatching. In A2, we aggregate neighbors with a fixed attention map as used in FSRL. In A3, we remove the embeddings of entities' own and encode them with only their neighbors. Experiments show that aggregating entity neighbors in an adaptive way and considering self-embedding can benefit the model performance.


(B) \textbf{Transformer Encoder Variants:} In B1, we replace the encoder by a concatenate operation on entity pairs as used in both GMatching and FSRL. In B2, we remove position embeddings in the Transformer encoder. Experiments indicate that the Transformer can effectively model few-shot relations, and position embeddings are also essential.


(C) \textbf{Matching Processor Variants:} In C1, we just obtain the embedding of reference set by averaging all reference representations. In C2, we only take the reference that is the most relevant to the query. In C3, we adopt the LSTM matching network as used in GMatching. Experiments indicate that our adaptive matching processor has superior capability in computing relevance between references and queries. 

\begin{table}[t]
	\centering\footnotesize\setlength{\tabcolsep}{3pt}
    \begin{tabular*}{0.48 \textwidth}{@{\extracolsep{\fill}}@{}l|l@{}}
    \toprule
    Tasks    & Head Entity: Obama   \\
    \midrule
    HasSpouse         & HasSpouse\_Inv, HasFamilyMember, BornIn \\
    Collaborate       & PoliticianOffice, Graduated, ProxyOf \\
    \midrule
                      & Head Entity: Microsoft \\
    \midrule
    ProxyFor          & ProxyOf, Leader, AgentControls \\
    CompeteWith       & Acquired, Products, Collaborate \\ 
    \bottomrule
    \end{tabular*}
    \caption{The most contributive relation neighbors in different tasks. Top 3 relation neighbors are shown.}
    \label{tab:NeighborAtts}
\end{table}

\begin{table}[t]
	\centering\footnotesize\setlength{\tabcolsep}{3pt}
    \begin{tabular*}{0.48 \textwidth}{@{\extracolsep{\fill}}@{}l|cc@{}}
    \toprule
    References   & Query~1 & Query~2 \\
    \midrule
    (Petersburg,    Virginia)   & .116 & .230 \\
    (Vacaville,     California) & .105 & .306 \\
    (Prague,        Czech)      & .107 & .314 \\
    (Cavaliers,     NBA)        & .208 & .072 \\ 
    (L.A. Lakers,   NBA)        & .464 & .078 \\
    \bottomrule
    \end{tabular*}
    \caption{Attention weights of 5-shot references, given two queries: Query~1 \texttt{\small (C. Bulls, NBA)} and Query~2 \texttt{\small (Astana, Kazakhstan)}. The task relation of all entity pairs is \texttt{\small SubPartOf}. 
    The references that are more related to the query achieve higher attention weights.
    }
    \label{tab:RelationAtts}
\end{table}

\subsection{Case Study for Adaptive Attentions}
To better understand the effects of adaptive attentions in the neighbor encoder and the matching processor, we conduct a case study. Table~\ref{tab:NeighborAtts} provides the most contributive relation neighbors with the highest attention weights in different tasks. We can see that the contributive neighbors for each entity in both tasks are different. The entities tend to focus more on the neighbors that are related to the task. Table~\ref{tab:RelationAtts} shows attention weights of references given different queries. The attention map of references is varied for each query, and the queries focus more on the related references. We can see that the attention weights are higher for location-related references when the query is location-related, while those are higher for organization-related references when the query is organization-related. This further indicates that our adaptive matching processor can aggregate references dynamically adaptive to the query, and benefits the matching process. All the above results further confirm our intuition described in the introduction. 

\subsection{Results on Different Relations}

\begin{table}[t]
    \centering\footnotesize\setlength{\tabcolsep}{5pt}
    \begin{tabular*}{0.48 \textwidth}{@{\extracolsep{\fill}}@{}lc|cccccc@{}}
    \toprule
    \multicolumn{2}{c}{} & \multicolumn{2}{c}{MRR} && \multicolumn{2}{c}{Hits@10} \\
    \cmidrule{3-4}
    \cmidrule{6-7}
    RId & \#~Candidate & MetaR & FAAN && MetaR & FAAN \\
    \midrule
    1  & 123     & .971  & \bf .974  && .971  & \bf .986 \\
    2  & 299     & .371  & \bf .533  && .453  & \bf .766 \\
    3  & 786     & .211  & \bf .352  && .524  & \bf .610 \\
    4  & 1084    & .552  & \bf .607  && .835  & \bf .846 \\
    5  & 2100    & .522  & \bf .595  && .643  & \bf .735 \\
    6  & 2160    & .216  & \bf .255  && .270  & \bf .336 \\
    7  & 2222    & \bf .153  & .112  && \bf .363  & .252 \\
    8  & 3174    & .292  & \bf .400  && .543  & \bf .697 \\
    9  & 5716    & .066  & \bf .084  && .133  & \bf .168 \\
    10 & 10569   & \bf .054  & .050  && .086  & \bf .128 \\
    11 & 11618   & \bf .082  & .013  && \bf .109  & .036 \\
    \bottomrule
    \end{tabular*}
    \caption{Results of MetaR and FAAN for each relation (RId) in NELL testing data. \#~Candidate denotes the number of candidate entities. {\bf Bold} numbers denote the best results of models.}
    \label{tab:ResultsDetails}
\end{table}

Besides the overall performance reported in the main results, we also conduct experiments to evaluate the performance of each task relation in NELL testing data. Table~\ref{tab:ResultsDetails} reports the results of the best baseline model MetaR and our model FAAN. According to the table, we find that the results of both models on different task relations are of high variance. The reason may be that the number of candidate entities is different, and the relations with large candidate set are usually hard to make predictions. Even so, our model FAAN has better performance in most cases, which indicates that our model is robust for different task relations.   

\section{Conclusion}

This paper proposes an adaptive attentional network for few-shot KG completion, termed as FAAN. Previous studies solve this problem by learning static representations of entities or references, ignoring their dynamic properties. FAAN proposes to encode entity pairs adaptively, and predict facts by adaptively matching references with queries. Experiments on two public datasets demonstrate that our model outperforms current state-of-art methods with different few-shot sizes. Our future work might consider other advanced methods to model few-shot relations, and exploiting more contextual information like textual description to enhance entity embeddings.

\section*{Acknowledgments}

We would like to thank all the anonymous reviewers for their insightful and valuable suggestions, which help to improve the quality of this paper. This work is supported by the National Key Research and Development Program of China (No.2017YFB0803305) and the National Natural Science Foundation of China (No.61772151). This work is also supported by Beijing Advanced Innovation Center of Big Data and Brain Computing, Beihang University.

\bibliographystyle{acl_natbib}
\bibliography{anthology,emnlp2020}

\end{document}